# Sensor Fusion of Camera and Cloud Digital Twin Information for Intelligent Vehicles

Yongkang Liu[1], *Student Member*, *IEEE*, Ziran Wang[2], *Member*, *IEEE*, Kyungtae Han[2], *Senior Member*, *IEEE*, Zhenyu Shou[3], Prashant Tiwari[2], and John H. L. Hansen[1], *Fellow*, *IEEE*

*Abstract* — With the rapid development of intelligent vehicles and Advanced Driving Assistance Systems (ADAS), a mixed level of human driver engagements is involved in the transportation system. Visual guidance for drivers is essential under this situation to prevent potential risks. To advance the development of visual guidance systems, we introduce a novel sensor fusion methodology, integrating camera image and Digital Twin knowledge from the cloud. Target vehicle bounding box is drawn and matched by combining results of object detector running on ego vehicle and position information from the cloud. The best matching result, with a 79.2% accuracy under 0.7 Intersection over Union (IoU) threshold, is obtained with depth image served as an additional feature source. Game engine-based simulation results also reveal that the visual guidance system could improve driving safety significantly cooperate with the cloud Digital Twin system.

## I. INTRODUCTION AND BACKGROUND

A report from the National Highway Traffic Safety Administration (NHTSA) revealed that 94% of road accidents are caused by human errors [1]. Therefore, a significant effort in developing intelligent vehicles has been made over the past decade, with the expectation that such technology could prevent accidents and improve efficiency. While there is great interest in migrating from human control to fully automated vehicles, factors such as system performance, manufacture cost, government regulation, and public safety suggest it is more likely that intelligent vehicles with different level of automation will coexist in large-scale traffic scenarios. In such a scenario, the automation levels of intelligent vehicles could range from level 0 (i.e., no automation) to level 5 (i.e., full automation) and has a different degree of human driver engagements [2]. Therefore, a visual guidance system is necessary to alert drivers with potential risks and/or provide lane level guidance, ensuring safe and efficient driving.

Object detection has been a critical part of both full automated driving systems and Advanced Driving Assistance Systems (ADAS), which determines whether objects of interested classes are present in an image, and identifies their sizes by using bounding boxes. It has been researched for decades in the field of computer vision, but only in recent years, the algorithm performance has been significantly increased with the development of Deep Convolutional Neural Networks (DCNNs) [3]. While almost all state-of-the-art approaches have been developed based on CNNs, it usually can be divided into two categories: two-stage detectors and one-stage detectors.

Two-stage detectors first propose a set of Region of Interests (RoIs), then categorize by separate classifier networks. Two-stage detectors have higher detection performance, but at the cost of high computation power and running time. A lot of efforts have been made to improve detection speed, including Fast R-CNN [4] and Faster R-CNN model [5]. One-stage detectors, on the other hand, skip the region proposal stage and use a single network to produce object detection locations as well as class prediction simultaneously. It tends to have fast inference time and low memory cost, which is well suited for real-time automated driving systems and ADAS. YOLO (You Only Look Once) has been attracting increasing attention along with its evolvement [6], [7]. It uses a pre-trained DCNN to extract features from the image, which has been split into grid cells and significantly reduces the resolution of input images. Other widely used algorithms, RetinaNet [8] and Single Shot Detector (SSD) [9], attempt to use DCNN's featured image pyramid [10] for efficient detection of objects of various sizes, trying to achieve the balance between speed and accuracy [11].

However, based solely on object detection sensors and the aforementioned computer vision techniques, only real-time information of the target vehicle can be perceived. Due to the short time horizon of detection, no historical data could be utilized to predict the behavior of the target vehicle. Digital Twin technology, as an emerging representation of Internet of Things (IoT), is able to provide cloud-based historical data to assist the prediction. Digital Twin often refers to systems with entities in the physical world and their digital replicas in the cyber world [12]. Although the Digital Twin concept has been widely studied and applied in the fields of aeronautics and space [12], [13], robotics [14], manufacturing [15], and informatics [16], it is still a relatively new concept in the automotive industry. Among the very few research studies of Digital Twin for vehicles, Alam and Saddik proposed a Digital Twin architecture reference model for the cloud-based ADAS [17], while Wang et al. conducted a field implementation of a Digital Twin-based ramp merging ADAS using three real passenger vehicles [18]. However, none of the existing studies proposes to fuse the cloud Digital Twin information with the camera information, aiming to better predict the behavior of the target vehicle and hence provide better guidance to the ego vehicle.

[1]Yongkang Liu and John H. L. Hansen are with the Electrical Department, University of Texas at Dallas, Richardson, TX 75080, USA (e-mail: {yongkang.liu, john.hansen}@utdallas.edu)

[2]Ziran Wang, Kyungtae Han and Prashant Tiwari are with Toyota Motor North America, InfoTech Labs, Mountain View, CA 94043, USA (e-mails: {ziran.wang, kyungtae.han, prashant.tiwari}@toyota.com)

[3]Zhenyu Shou is with the Department of Civil Engineering and Engineering Mechanics, Columbia University, New York, NY 10027, USA (e-mail: zs2295@columbia.edu)

The rest of the paper is organized as follows: Section II delivers the problem statement of this study. Section III introduces the proposed sensor fusion methodology. Simulation design in the Unity game engine and the evaluation of its results are included in Section IV. Section V concludes the paper with potential future directions.

## II. PROBLEM STATEMENT

Most of the existing automated driving systems or ADAS rely on perception sensors such as cameras to perceive the real-time surrounding information, and make prediction regarding the future behaviors of other road entities (e.g., target vehicle). However, the perceived information only includes their dynamics status (e.g., vehicle speed, position) in real-time (or at most a short period of time), without a long horizon of historical data.

By leveraging Digital Twin technology, virtual copies of real transportation entities (e.g., vehicles, drivers, pedestrians) are created on the cloud, where the data produced by these entities can be stored and processed by the cloud-based processing modules/algorithms. Once the online process is completed, the predicted information of the target entities (e.g., target vehicle's intention to change lane) by the cloud server can augment the perceived information by ego vehicle's cameras, assisting the automated driving systems or ADAS to make better decisions.

However, a key issue arises as how to correctly overlay the cloud Digital Twin information received through vehicle-to-cloud communication onto the correct target vehicle. Equally speaking, from the ego vehicle's perspective, how to identify which is the target vehicle whose information has been shared [19]. It is true that cloud Digital Twin information includes a rich set of parameters regarding the target vehicle (both historical and real-time data), however, cameras can only detect a limited amount of real-time data. To the best of the authors' knowledge, how to correctly fuse two different data sources, so the historical big data on the cloud can enhance real-time perception data remains an unsolved question in the automotive domain.

In this study, we propose a sensor fusion technology to leverage the camera information and cloud Digital Twin information, aiming to predict the lane change behavior of other vehicles. Specifically, position information measured by vehicles' Global Navigation Satellite System (GNSS) is utilized to identify the target vehicle (with a potential to change lane), hence the correct advisory can be visualized to the driver as a feature of ADAS on intelligent vehicles. The major contributions of our study are listed as follows.

- To the best of the authors' knowledge, this is one of the first studies that visualizes cloud Digital Twin information to assist the decision making of intelligent vehicles.
- The difference between using one camera source (RGB camera) and using two camera sources (RGB and depth cameras) for target vehicle identification is studied.
- Human-in-the-loop simulation is conducted in a game engine-based intelligent vehicle simulation environment, where the safety benefits of implementing the proposed sensor fusion methodology to the lane change scenario are shown.

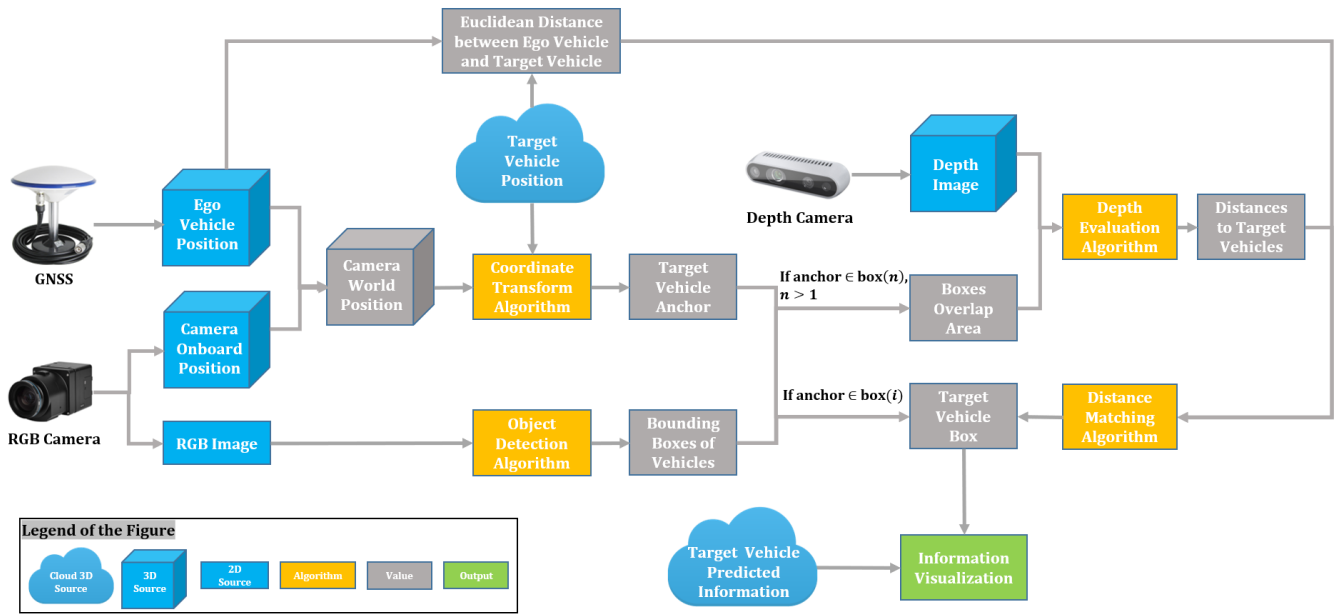

Fig. 1. System architecture of the proposed sensor fusion methodology

## III. Proposed Sensor Fusion Methodology

In this study, we assume all vehicles in the traffic environment have internet access, so their Digital Twins can be built on the cloud server based on the data they transmitted with the cloud (e.g., speed, position, etc.). Specifically, position information can be measured by vehicles' onboard GNSS, and be frequently updated to the cloud Digital Twin. Since the focus of this study is the sensor fusion of GNSS and camera, we further assume Digital Twin model on the cloud server can provide useful information (driver type in this study) based on both historical and real-time data, so it can be visualized and assist the decision making of the intelligent vehicle.

The system architecture of the proposed methodology can be illustrated as Fig. 1. There are four different data sources of the ego vehicle, including three 3D sources (i.e., ego vehicle position, camera onboard position, and depth image) and one 2D source (i.e., RGB image). Regarding the target vehicle, all data comes from the cloud server, where we utilize the vehicle position for sensor fusion, and the predicted information (i.e., driver type) for the final visualization of ADAS.

We propose four key algorithms in this architecture to tackle the sensor fusion problem, where their objectives and technical details are introduced in this section.

### A. Coordinate Transform Algorithm

To transform a 3D GNSS coordinate to a 2D image point, the pinhole camera projection model is used here and two transformation matrices are required. They are the parameters used in a camera model to describe the relationship between the 3D coordinate of a point in world referenced frame and the 2D coordinate of its projection onto the image plane [20], as shown in Fig. 2.

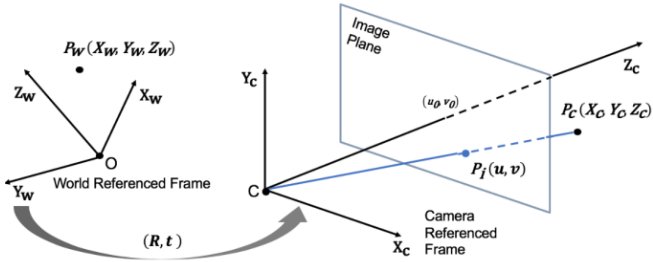

Fig. 2. Transformation of the world point to image plane

The extrinsic camera parameter matrix, also known as extrinsic parameters, are the parameters used to identify the transformation between the camera referenced frame and the world referenced frame. It consists of a 3x3 rotation matrix $R$, which brings the corresponding axes of the two frames into alignment. Followed by a translation vector $t$, which describes the relative positions of the origins of the two reference frames.

Therefore, given a 3D GNSS point $P$ in world referenced frame $P_w$ ($X_w, Y_w, Z_w$), the relationship between $P_w$ and the corresponding point $P_c$ ($X_c, Y_c, Z_c$) in camera referenced frame is

$$P_w = [R|t]P_c \quad (1)$$

The intrinsic camera parameter matrix, or intrinsic parameters, are the parameters of the camera itself, such as the focal length, lens distortion and the transformation between image plane coordinates and pixel coordinates. Also illustrated in Fig. 2, let ($u_0, v_0$) be the coordinates of the principal point in pixels, $d_x, d_y$ are the physical size of pixels, and $f$ is the focal length. The matrix containing the intrinsic parameters is

$$M_i = \begin{bmatrix} Z_c d_x/f & 0 & -Z_c d_x u_0/f \\ 0 & Z_c d_y/f & -Z_c d_y v_0/f \\ 0 & 0 & Z_c \end{bmatrix} \quad (2)$$

The coordinates ($u, v$) of our GNSS point in the image plane $P_i$ are

$$P_i = M_i^{-1} P_c \quad (3)$$

This point will be served as an "anchor point", which helps the identification of the target vehicle in the future matching algorithm.

### B. Object Detection Algorithm

For the development of ADAS, both accuracy and computational cost are essential factors. The detection module needs to be reliable, but also needs to operate at high speed to allow sufficient processing time for the rest modules to take actions. Under this consideration, one-stage detector is adopted in our approach.

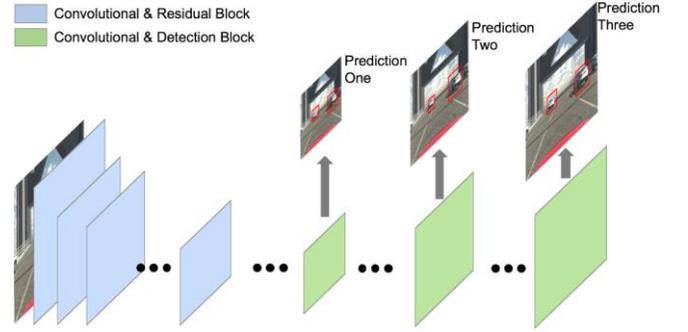

Fig. 3. Object detection network architecture

We build our object detection module based on PyTorch YOLOv3 implementation [21] pre-trained on the COCO [22] dataset, where the network architecture is illustrated in Fig. 3. The original network draws a bounding box with a class id and confidence score for each detected object. However, in our approach, we replace the confidence score with distance calculated from depth image, and all bounding boxes information is cached for the matching step.

### C. Depth Evaluation Algorithm

Additional features might be necessary if detected bounding boxes are not perfectly isolated (namely anchor point is located in multiple detected bounding boxes), as shown in Fig. 4 (a). Here we propose to use spatial information to improve matching accuracy. For example, Fig. 4 (b) is the depth image taken at the same time, the distance from detected vehicles to the camera can be determined using *Algorithm 1*. For each detected bounding box, a total number of $n$ points are randomly sampled to calculate the distance. To ensure sampled points have good representativeness, first the box size is decreased by a certain threshold, then the points are selected from the lower quarter area of the box. This distance set will be served as supplementary information in our matching algorithm.

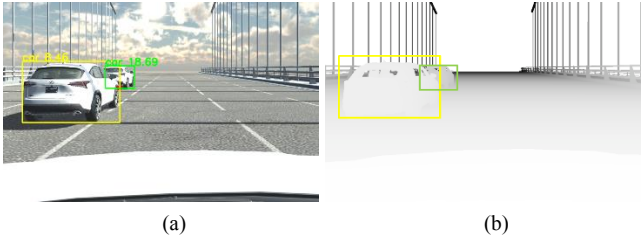

Fig. 4. (a) One scenario that detected bounding boxes are not perfectly isolated, numbers are the distance results from depth evaluation algorithm. (b) Depth image taken at the same time, used as the input of depth evaluation algorithm

---

*Algorithm 1: Depth Evaluation Algorithm*

**Input**: Depth image ($Img_d$), detected bounding boxes ($B$), box resize threshold ($th$), total number of sample points ($n$).
**Output**: Distance set ($D$) from detected vehicles to camera.
1: **for** each detected bounding box $B_i \in B$
2:    decrease the box size according to the box resize threshold ($th$), select the lower ¼ area of the box as $A$;
3:    **for** each point $p_j$, $j < n$
4:      randomly select the point position where $p_j \in A$;
5:      calculate the distance ($pd_j$) between $p_j$ and camera using value from $Img_d$;
6:      put $pd_j$ into the temporary distance set $\Theta$;
7:    **end for**
8:    calculate the distance $d_i = avg(\Theta)$;
9:    put $d_j$ into the distance set $D$;
10: **end for**
11: **return** $D$

---

### D. Distance Matching Algorithm

The object detection algorithm will have multiple detection results if a number of vehicles are presented in a single image. However, since the proposed ADAS needs to correctly overlay the right information to its associated target vehicle, we propose a distance matching algorithm in *Algorithm 2*.

We first introduce the baseline matching approach based on the anchor point calculated from coordinate transformation. If the anchor point is located exclusively inside of one detected bounding box, this box will be selected as our target vehicle box. This approach yields a satisfied result most of the time, but fails to distinguish the target vehicle in some tricky scenarios as we introduced in the previous subsection. Therefore, spatial knowledge as additional features are employed. The distance set acquired from the depth evaluation algorithm will be used to compare with the distance directly obtained from GNSS. The box with minimum distance difference is selected as the target vehicle box.

---

*Algorithm 2: Distance Matching Algorithm*

**Input**: Anchor point ($P_i$), detected bounding boxes ($B$), Distance set ($D$) from detected vehicles to camera, distance directly obtained from GNSS ($D_g$).
**Output**: Target vehicle bounding box ($B_t$).
1: **for** each detected bounding box $B_n \in B$
2:   **if** $P_i \in B_n$ **then**
3:     put $B_n$ into the temporary set $\Theta$;
4:     $counter$ += 1;
5:   **end if**
6: **end for**
7: **if** $counter == 1$ **then**
8:   **return** $B_t = \Theta$
9: **else**
10:   **for** each box $B_j \in \Theta$ and distance $d_j \in D$
11:     calculate the distance difference $\Delta d_j = d_j - D_g$;
12:     put $\Delta d_j$ into the temporary set $T$;
13:   **end for**
14:   find the index $i \in len(T)$ with min(distance difference);
15:   set target box $B_t = \Theta_i$;
16:   **return** $B_t$
17: **end if**

---

## IV. GAME ENGINE SIMULATION AND RESULTS EVALUATION

### A. Game Engine Simulation in Unity

Naturalistic driving data and real-world testbeds are essential to modeling driver behaviors [23], [24], but the data collection and system implementation process are costly and time-consuming. Game engine-based simulation testbed, on the other hand, is helpful at the proof-of-concept stage with the ability to easily visualize cooperative driving systems that involve a large number of vehicles, and the flexibility to acquire extensive data under different and sometimes dangerous driving scenarios.

Game engines (such as Unity [25] and Unreal [26]) are software systems that consist of a rendering engine for 2-D or 3-D graphics, a physical engine for collision detection and response, and a scene graph for the managing elements (e.g., models, sound, scripts, threads, etc.) [18]. Different autonomous driving test environments have been built by game engines due to their advantages of visualization and integration, where the most widely used examples are LGSVL [27] and CARLA [28]. Some small-scale game environments were also developed by various studies to prototype connected vehicles [29], ADAS [30], and autonomous vehicles [31].

In this study, Unity game engine is utilized to demonstrate and evaluate our proposed sensor fusion methodology. Both RGB and depth cameras are implemented on the ego testing vehicle, which can generate real-time RGB images (such as Fig. 4 (a)) and depth images (such as Fig. 4 (b)). GNSS module is also implemented to enable vehicles to get real-time positions (as a 3D coordinate in the game environment). Since we assume all vehicles are intelligent vehicles with vehicle-to-cloud communication (i.e., internet of vehicles), the ego vehicle has access to target vehicle predicted information in the simulation environment. Augmented Reality (AR)-based information visualization method is adopted to display the cloud-based predicted information to the ego vehicle.

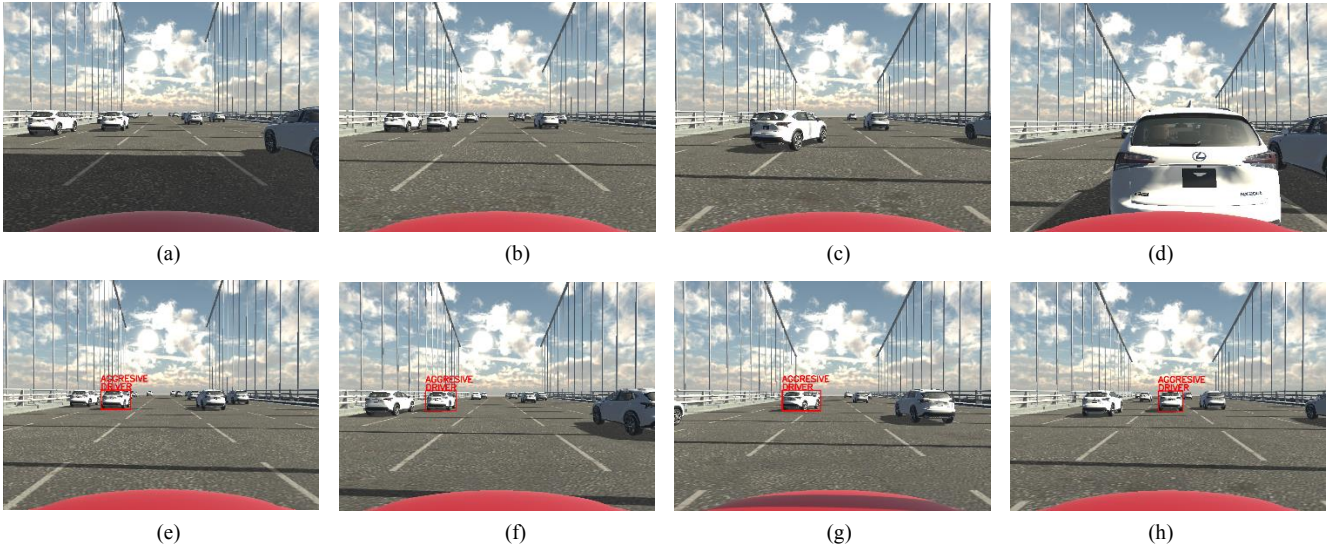

Fig. 5. Human-in-the-loop simulation in game engine with baseline approach (a)-(d), and sensor fusion-based ADAS approach (e)-(h)

*B. Distance Matching Algorithm Evaluation*

The first evaluation is a post-processing assessment to examine the performance of the distance matching algorithm. Various of tricky scenarios (e.g., surrounding vehicles are very close to each other from the ego vehicle's perspective) are generated in the simulation environment. The proposed distance matching approach (fusion of RGB camera and depth camera) is compared with the baseline approach (RGB camera only).

As shown in Fig. 6, results are evaluated with the target vehicle identification accuracy and Intersection over Union (IoU). It is clearly observed that the distance matching approach outperforms the baseline approach in terms of the target vehicle detection accuracy. By introducing spatial knowledge from depth images as additional features, a 7.7% improvement (79.2% over 73.5%) is obtained when the IoU threshold is set to 0.7. One thing to note is that among the failed matching cases, most of them are caused by the detection failure of overlapped vehicles. By introducing the depth camera, many of those tricky cases are successfully detected, and the correct target vehicle is picked out.

*C. Human-in-the-Loop Simulation Evaluation*

In the second experiment, a human-in-the-loop simulation is conducted to evaluate the effectiveness of our sensor fusion-based ADAS. A real-world multi-lane highway scenario is built in Unity game engine, where the ego vehicle is surrounded by neighbor vehicles that might suddenly affect the ego vehicle's trajectory. Participants of this simulation are asked to drive the ego vehicle to stay in its own lane and maintain safety margin with others.

Fig. 5 (a) – (d) shows the snapshots of one simulation (out of many trials) with the baseline approach, where the participants drive the ego vehicle with no additional information. Since the left neighbor vehicle suddenly changes the lane to the right, the ego vehicle has limited time to react, hence it almost collides with that neighbor vehicle. On the other hand, Fig. 5 (e) – (h) shows the simulation snapshots with the proposed sensor fusion-based ADAS approach, where AR-based visualization identifies the target vehicle and displays the predicted information from the cloud (i.e., that vehicle has an aggressive driver).

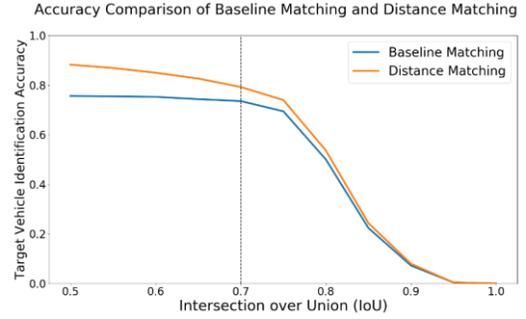

Fig. 6. Target vehicle identification accuracy-IoU curve

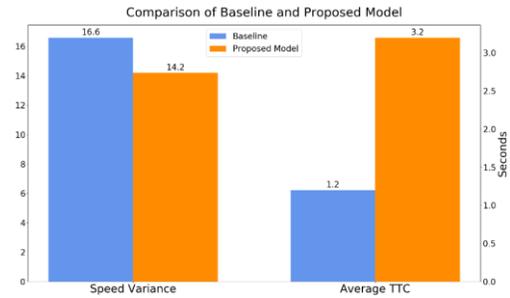

Fig. 7. Comparison results from simulation

We compare the aforementioned approaches from the driving safety perspective, i.e., compare the values of speed variance and time-to-collision (TTC). The results are shown in Fig. 7, where the speed variance is decrease from 16.6 to 14.2 by implementing the proposed model. Generally, the accidents rates increase with increased speed variance for all classes of roads, so a reduced speed variance value proves the safety of our model. Additionally, a significant increase of average TTC is observed from 1.2 seconds to 3.2 seconds. A larger TTC means the driver will have more time to react if any emergency happens. In conclusion, both speed variance and TTC results validate the safety benefit of the proposed sensor fusion-based ADAS. The participant of the simulation is notified by this information, and is able to decelerate ahead of time as a precaution to the potential dangerous behavior (i.e., sudden lane change) of the target vehicle.

## V. Conclusions and Future Work

In this study, a sensor fusion approach aiming to visualize cloud Digital Twin information is proposed. A target vehicle identification strategy utilizing spatial information is explored, achieving a 79.2% accuracy under 0.7 IoU threshold. Human-in-the-loop simulation in Unity game engine is conducted, revealing the huge safety benefits (in terms of speed variance and TTC) of implementing the proposed sensor fusion methodology to intelligent vehicles.

Since the transmission of GNSS information may have delays and limited update frequency, one of the future research directions is to integrate object tracking algorithm into the system to maintain the target vehicle identification accuracy. Another potential direction is to implement the proposed system from simulation to real-world passenger vehicles, with a separate monitor display or an advanced design of head-up display (HUD).

## Acknowledgment

The contents of this paper only reflect the views of the authors, who are responsible for the facts and the accuracy of the data presented herein. The contents do not necessarily reflect the official views of Toyota Motor North America.